\documentclass{article}

\usepackage[dblblindworkshop, final]{neurips_2026}
\workshoptitle{Who Verifies the Agents? Toward Reliable Agent Development}
\setcitestyle{numbers,square,comma}

\usepackage[utf8]{inputenc}
\usepackage[T1]{fontenc}
\usepackage{url}
\usepackage{booktabs}
\usepackage{array}
\usepackage{amsfonts}
\usepackage{amsmath}
\usepackage{amssymb}
\usepackage{microtype}
\usepackage{xcolor}
\usepackage{colortbl}
\usepackage{graphicx}
\usepackage{algorithm}
\usepackage{algorithmic}
\usepackage{hyperref}

\definecolor{vind}{HTML}{4F46E5}
\definecolor{vinddk}{HTML}{3730A3}
\definecolor{vindbg}{HTML}{EEF2FF}
\definecolor{aamb}{HTML}{D97706}
\definecolor{ateal}{HTML}{0D9488}
\definecolor{frose}{HTML}{BE123C}
\definecolor{inkmut}{HTML}{64748B}

\hypersetup{
  colorlinks=true,
  linkcolor=vinddk,
  citecolor=vinddk,
  urlcolor=vinddk,
  breaklinks=true,
}

\newcommand{\gon}{{\color{ateal}$\checkmark$}}
\newcommand{\goff}{{\color{frose}\textbf{\textsf{x}}}}
\newcommand{\pc}{\cellcolor{vindbg}}

\title{Who Verifies the Verifier? Co-Evolving\\ Inspectable Graders with Self-Improving Agents}

\author{
  Xing Zhang$^{1}$\quad
  Guanghui Wang$^{1}$\quad
  Yanwei Cui$^{1}$\quad
  Ziyuan Li$^{2}$\\
  \textbf{Wei Qiu$^{2}$\quad
  Bing Zhu$^{2}$\quad
  Peiyang He$^{1}$\thanks{Corresponding author: \texttt{peiyan@amazon.com}}}\\
  $^{1}$AWS Forward Deployed Engineering\\
  $^{2}$HSBC Holdings Plc., HSBC Technology Center, China
}

\begin{document}

\maketitle

\begin{abstract}
We changed the agent: did it actually get better? Every self-improving agent loop answers this hundreds of times, and every answer comes from a verifier. On open-ended tasks none exists, so the loop is handed a hand-written rubric or a bare LLM judge grading output from a model like itself, inviting reward hacking and shared blind spots. We make the verifier the evolving object: an inspectable expression over small, mostly deterministic drawback detectors, synthesized from clustered failures, gated at birth, and selected for agreement with a ten-item anchored reference set plus consensus over unlabeled outputs, never for the agent's score. On MBPP+ it gains $+0.21$ held-out agreement over the hand-authored seed composition, on every seed, and ends ahead of the bare LLM judge it contains. One finding should change how co-evolved verifiers are validated: removing the anchor guards collapses the verifier into a vacuous always-pass grader, yet that collapsed verifier trains skills \emph{just as well}. Downstream task score cannot certify a self-evolved verifier. Score does answer sufficiency, and there an evolved verifier can substitute: \emph{Double Ratchet}, pairing the verifier with a lifecycle-managed skill loop, retains 88--110\% of the lift that ground truth or a rubric buys the same loop, across code generation, enterprise text-to-SQL, and reference-free report generation. When evolved skills gamed the report rubric, an outer judge caught it and one added detector repaired it; the judge itself was wrong until given the task contract.
\end{abstract}

\section{Introduction}

We update an agent's prompt, add a tool, or change its reasoning strategy: did it actually get better? A self-improving agent asks this every round. The standard recipe wraps a frozen language model in a loop that writes skills, retrieves them, measures their contribution, and retires the ones that hurt \citep{zhang2026ratchet,wang2023voyager,shinn2023reflexion}, and it consumes one expensive ingredient per round: a verifier that says whether an attempt passed. On code benchmarks that ingredient is free, since unit tests exist. In deployed systems it is the bottleneck, and the industry's own products document the workaround: Claude's managed agents have developers hand-write outcome rubrics for a separate grader \citep{anthropic2026outcomes}, and Codex's Record \& Replay has a human demonstrate each workflow once \citep{openai2026recordreplay}. Both are manual substitutes for a missing verifier, supplied one rubric or demonstration at a time.

This paper automates that step: \emph{can an agent's verifier be evolved from a handful of anchored examples, drive learning as well as ground truth, and not be gamed in the process?} Two clarifications determine how every result below should be read. First, \textbf{the verifier, or grader, is the function that grades one output}, and it is the only thing that evolves; the task sets and their references are fixed before any loop starts, so no result here can come from an easier test. Second, in our target setting \emph{no accurate verifier exists}, so there is no baseline verifier to beat: the yardstick is how much of the improvement an accurate verifier \emph{would have enabled} a system recovers without one. Matching a ground-truth-driven reference loop is the win condition, not a consolation.

\begin{figure*}[t]
\centering
\includegraphics[width=\textwidth]{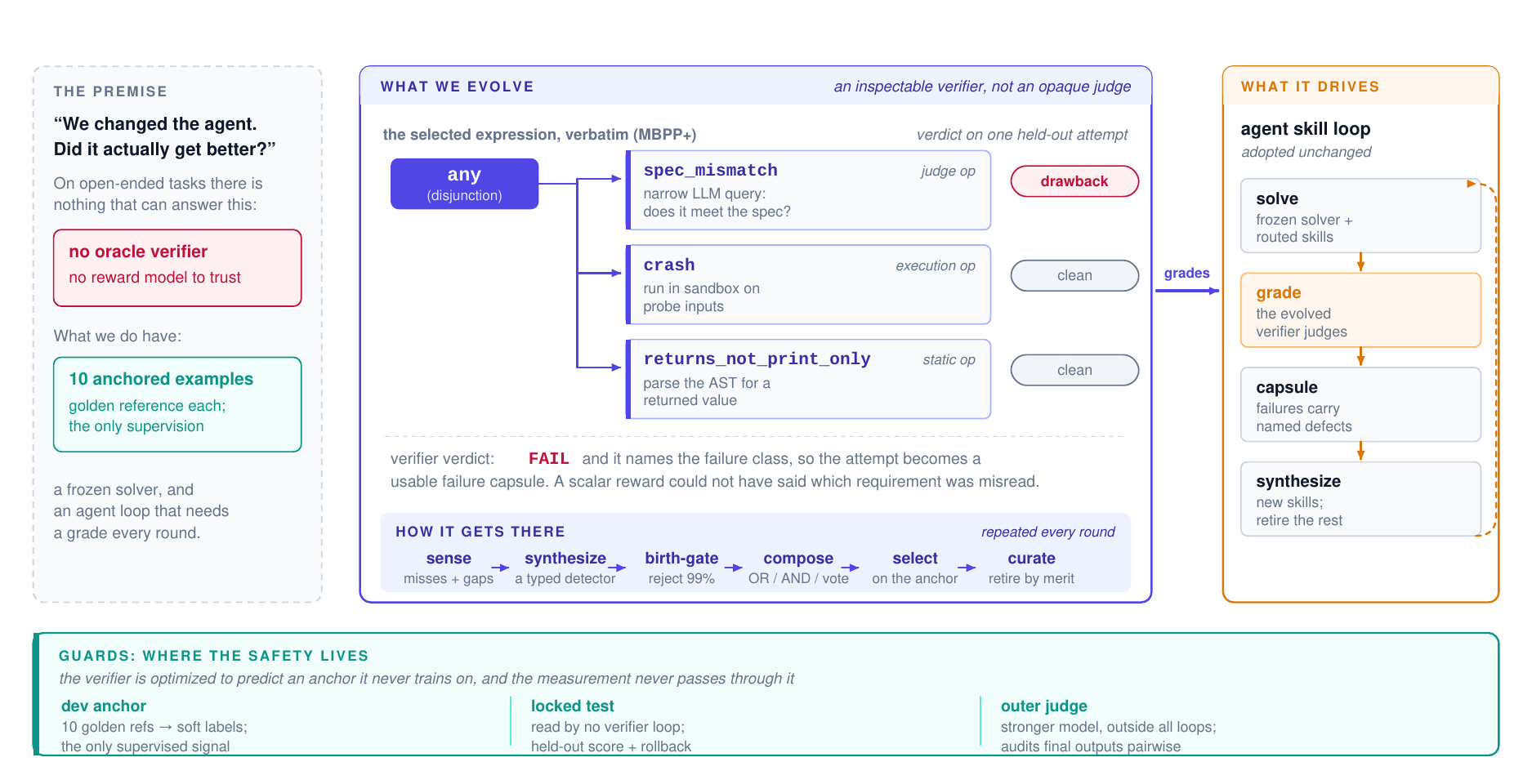}
\caption{What evolves. \textbf{Center}: the verifier selected on MBPP+, verbatim, as an expression over typed detectors spanning the three cost tiers, with each leaf's verdict on one held-out attempt. Naming the failure class rather than emitting a scalar is what turns a graded attempt into a usable failure capsule. \textbf{Left}: the premise, no oracle verifier and ten anchored examples. \textbf{Right}: the consumer, a lifecycle-managed skill loop \citep{zhang2026ratchet} adopted unchanged, whose training the evolved verifier grades. \textbf{Bottom}: the anchor guards.}
\label{fig:hero}
\end{figure*}

The design rests on one stance about verification. We rarely know what \emph{good} is, but given an output we can usually find \emph{drawbacks}, so a clean verdict means no known drawback was found rather than certified correctness. That framing is what makes bootstrapping possible, and it determines the shape of the artifact (Figure~\ref{fig:hero}): the verifier is an expression over a pool of small typed detectors, each checking one failure class, synthesized from clustered failures, gated at birth, scored by marginal contribution, and retired when useless. Detectors span the three cost tiers a real deployment already has (static checks, execution against a sandbox or a live warehouse, and narrow judge queries), so composing them folds heterogeneous signals into one grader instead of trusting a single scalar. Being mostly deterministic, they fail differently from the model being graded, which is what resists the shared-blind-spot collusion a bare LLM judge invites; being a readable expression, the grader localizes a failure to a nameable check that can be repaired by hand (\S\ref{sec:hack}).

A verifier that grades the loop that produced it is an obvious Goodhart hazard \citep{goodhart1984problems,gao2023scaling}, so anchor discipline is strict and load-bearing. Selection never asks whether the agent scores well: it asks whether the verifier agrees with a tiny anchored reference set, the only supervised signal verifier evolution reads, and with detector consensus on a large unlabeled set. A separate locked set is read by no verifier loop and serves only for verifier evaluation and skill-side rollback. We test by ablation whether that discipline carries the safety load, and the answer bears directly on how this community should validate co-evolved verifiers at all.

Our contributions are as follows.
\begin{itemize}\itemsep1pt
\item \textbf{Verifier evolution, with an objective that needs no downstream metric.} We formulate the grading function, rather than the artifact being graded, as the evolving object: anchored search over compositions of atomic drawback detectors under a lifecycle of synthesis, birth gating, marginal-contribution scoring, and retirement. Selection is recall-weighted agreement with a ten-item anchor times reliability-weighted consensus over unlabeled outputs, and the result beats both the hand-authored composition it starts from and the bare LLM judge it contains.
\item \textbf{Where the safety lives, and why task score cannot find it.} Anchor discipline, not the detector lifecycle, is load-bearing for verifiers, inverting the lesson from skill evolution. Since a collapsed verifier trains skills as well as a valid one, \emph{validating a self-evolved verifier by downstream performance is unsound}: the most natural validation experiment would certify a grader that passes everything.
\item \textbf{Double Ratchet, and a reward-hacking episode end to end.} Co-evolving the verifier with a skill loop retains 88--110\% of the lift ground truth buys, across three task areas. On the reference-free task we document proxy gaming caught by an outer judge, repaired by one added detector, and an audit of the auditor showing the outer judge needs the task contract.
\end{itemize}

\section{Related work}

\paragraph{Self-evolving agents assume the verifier.}
Agents grow skill libraries, revise from failure traces, and bank reusable experience as insights, manuals, memories, or workflows \citep{wang2023voyager,shinn2023reflexion,madaan2023selfrefine,zhao2024expel,chen2024automanual,park2023generative,wang2024awm}; a recent wave evolves, analyzes, and benchmarks such libraries \citep{wu2025evolver,yang2026autoskill,xia2026skillrl,alibaba2026trace2skill,wang2026strategygenes,zhang2026ecs,li2026skillsbench}, and two surveys map the landscape \citep{zhang2025surveyselfevolving,wang2024survey}. Ratchet \citep{zhang2026ratchet,zhang2026librarydrift} showed that unmanaged LLM-authored libraries drift and that a minimal lifecycle restores the gains; that published loop is adopted unchanged here as the skill-side component and is not a contribution of this work. Other lines search over agent code \citep{hu2025automated}, inherit the architecture-search tradition \citep{elsken2019nas}, or optimize prompts and pipelines against a metric \citep{khattab2024dspy,yang2023opro,fernando2024promptbreeder,yuksekgonul2024textgrad}. All of them optimize \emph{against} a verification signal taken as given. We evolve that signal, and ask which part of the discipline carries the safety load once the grader itself moves.

\paragraph{Searched objectives are scored by a metric that already exists.}
Existing work evolves loss functions \citep{liu2021cseautoloss}, whole learning algorithms \citep{real2020automlzero}, symbolic pruning metrics \citep{dong2024prunerzero}, and program code \citep{novikov2025alphaevolve} against supplied evaluators. The search space is analogous, primitive checks composed into a program, but the fitness differs in kind: each is scored by a metric taken as valid, with AlphaEvolve stating outright that progress must be machine-gradable, the precondition our regime lacks. We supply the missing foothold: anchored agreement plus consensus regularization in place of a downstream score.

\paragraph{Learned graders trade the assumption for opacity.}
Reward design is delicate even in classical RL \citep{ng1999pbrs}; RLHF reward models are over-optimized by the policies they train and misspecification produces systematic gaming \citep{ouyang2022rlhf,gao2023scaling,pan2022effects,skalse2022defining,amodei2016concrete}; AI-feedback and self-rewarding setups let the model judge itself \citep{bai2022constitutional,yuan2024selfrewarding}; and LLM judges are widely audited for bias \citep{zheng2023judging,dubois2024alpacafarm,gu2024surveyllmjudge,chen2024alignment}. We share the reward-hacking hazard but not the structure: our grader is a composition of typed detectors optimized to predict an anchor it never trains on, so it can be read and repaired by hand, and its agent-facing use is confined to training. Agent-as-judge co-evolution \citep{zhuge2024agentjudge} is closest in spirit and validates the judge through the system it grades, which \S\ref{sec:guards} shows a vacuous grader also passes. The evaluators nearest our tasks fix in advance what we synthesize, gate, and retire: CodeScore \citep{dong2024codescore} predicts functional correctness from the execution-labeled supervision we take to be missing, and SQLHD \citep{yang2025sqlhd} needs no ground truth but fixes a hand-authored family of metamorphic relations \citep{chen1998metamorphic}. Either could be one typed detector in our pool, so we read them as components rather than competitors, while agentic text-to-SQL pipelines \citep{deng2025reforce} improve the solver, an orthogonal axis. We use MBPP+ \citep{austin2021mbpp,liu2023evalplus} and Spider~2.0 \citep{lei2025spider2} as hard anchors for certifying evolved verifiers, in the execution-graded tradition of HumanEval \citep{chen2021codex}, not as leaderboard entries.

\section{The verifier and its evolution}

Let a task family come with tasks $t$ and a frozen solver mapping $t$ to an output $y$. A verifier $M$ maps $(t,y)$ to a verdict in $\{\textsc{pass},\textsc{fail},\textsc{abstain}\}$, and $M$ is what we search over, with tasks and references held fixed. Three disjoint splits play the usual roles with one twist each. \textbf{train} is large and unlabeled: solver outputs here expose where candidate verifiers disagree or abstain, and in co-evolution the evolved verifier grades these tasks for the skill loop. \textbf{dev} is tiny and anchored, ten items throughout: each task carries a golden reference against which a teacher emits soft pass/fail labels, the only supervised signal verifier evolution reads. Those references grade only the ten tasks that carry them, while the evolved verifier must grade arbitrary outputs on tasks that have none. Dev is deliberately small because sparse anchoring is the realistic regime: fresh labels in deployed systems arrive as a trickle of demonstrations and incident reviews. \textbf{test} is locked against the verifier: it holds the strongest available reference (unit tests, official execution comparison, the report rubric) and no part of verifier evolution reads it, so every validity number below is fully out of sample. The skill loop alone touches it, for held-out scoring and rollback, as specified by Ratchet \citep{zhang2026ratchet} and kept unchanged because a deployed agent rolls back against the cases it is judged on. The reference and co loops roll back identically, so the retention comparison that carries our claim stays clean, and rollback fired in only 23 of their 2{,}900 rounds, suggesting limited direct exposure; absolute skill-side levels are nonetheless optimistic against a never-touched third split. The naive ablation switches rollback off, so its skill side is less protected, not more.

\subsection{Verifiers as compositions of drawback detectors}
\label{sec:ops}

An \emph{op} is an atomic drawback detector $o(t,y,c) \rightarrow \{\textsc{drawback},\textsc{clean},\textsc{abstain}\}$ with $c$ a shared context, checking exactly one failure class and abstaining otherwise. Each task family starts from a handful covering its obvious failure modes, cheap to write by hand or have an LLM draft from the task spec, and the lifecycle grows the pool from there. Ops come in three cost tiers: static ops parse and inspect the artifact; execution ops run it, against a sandbox for code or a live warehouse for SQL, so the environment itself supplies part of the ground truth; judge ops ask an LLM one narrow question. A fixed non-evolving \emph{root} per family validates basic structure and caches the parsed artifact the other ops read.

A \emph{verifier expression} composes op verdicts with logical operators: a node is a leaf op, a disjunction (a drawback if any child finds one), a conjunction (only if all children agree), a negation, an unweighted $K$-of-$k$ vote, or a weighted vote against a threshold (Appendix~\ref{app:lang} gives the grammar). Abstaining children are excluded from their combinator, a node whose children all abstain abstains itself, and the output passes when the fixed root conjoined with the evolved expression finds no drawback. Ops are pure functions of $(t,y)$, so verdicts are cached and evaluation reduces to table lookups. Negation and weighted votes stay available every round but appear in none of the selected verifiers, so every expression we report reads as a plain drawback disjunction or vote, fully reproducible from its expression string plus the registered op pool.

\subsection{The verifier loop}

Algorithm~\ref{alg:loop} sketches one round. Two signals feed synthesis: \emph{misses}, dev items the incumbent passes although their soft label is fail, and \emph{gaps}, train outputs on which the pool abstains entirely or splits. Both are clustered by failure class under a closed task-specific taxonomy, and a synthesizer authors a typed op spec for any recurring cluster, admitted only if it passes the birth gate. Ops born from anchored misses enter \emph{active} and are selectable, their birth evidence already tying them to the anchor; ops born only from unlabeled gaps enter as \emph{shadows}, recording verdicts every round but unselectable until their verdicts demonstrably raise dev agreement, since nothing yet shows they track quality rather than consensus noise. Candidates then come from an LLM composer that sees the op catalog, the current gaps, and an elite history of the best expressions with their shortcomings, plus mutations and crossovers of those elites. Selection keeps the best-so-far candidate by
\begin{equation}
S(e) \;=\; A_{\mathrm{dev}}(e) \cdot A_{\mathrm{train}}(e)^{w} \;-\; \lambda\, C(e),
\label{eq:score}
\end{equation}
where $A_{\mathrm{dev}}(e)$ is agreement with the dev soft labels, $A_{\mathrm{train}}(e)$ agreement with the consensus verdict of the opining pool ops on unlabeled train outputs, $C(e)$ expression size, and $w,\lambda$ small constants. Both agreement terms are weighted, and each weighting is a safety choice rather than a tuning knob (constants in Appendix~\ref{app:lang}). \emph{Dev agreement is recall-weighted}, a missed drawback costing twice a false alarm, so no candidate reaches a perfect score by acing the easier class alone. \emph{Consensus is reliability-weighted} by each op's anchored leave-one-out marginal, so anchor-proven ops lead the vote; otherwise a bloc of correlated detectors, cheap to synthesize and liable to share a blind spot, would define the very pseudo-label the regularizer rewards agreeing with. $A_{\mathrm{train}}$ is thus a transfer regularizer pulled toward the anchor rather than a popularity vote, and entering multiplicatively against a fail-closed $A_{\mathrm{dev}}$ it never selects a verifier on its own: a verifier that memorizes ten dev items but behaves erratically loses, and a degenerate always-fail verifier loses on dev. Two guards apply: selection fails closed, so a candidate with no usable dev opinion is unselectable, and a validity gate drops candidates that pass, fail, or abstain on everything. Per-op fitness is the leave-one-out marginal of the incumbent expression, and non-positive ops retire after a grace period.

\begin{algorithm}[t]
\caption{One round of verifier evolution}
\label{alg:loop}
\begin{algorithmic}[1]
\STATE {\color{vinddk}\textbf{Sense}} \ {\footnotesize\color{inkmut}(where is the verifier wrong?)}
\STATE \quad misses $\gets$ dev items the incumbent passes but the soft label fails
\STATE \quad gaps $\gets$ train outputs where the pool abstains or splits
\STATE {\color{vinddk}\textbf{Grow}} \ {\footnotesize\color{inkmut}(expand the detector pool)}
\STATE \quad cluster misses $\cup$ gaps by failure class; author one typed op per recurring cluster
\STATE \quad birth gate: fires on $\geq$ half its cluster and clean on known-good $\Rightarrow$ admit (active if anchored, else shadow)
\STATE {\color{vinddk}\textbf{Select}} \ {\footnotesize\color{inkmut}(recombine, keep the best)}
\STATE \quad candidates $\gets$ LLM compositions over the pool, plus mutations and crossovers of elites
\STATE \quad keep $\arg\max_e S(e)$ over candidates and the incumbent, behind the validity gate (Eq.~\ref{eq:score})
\STATE {\color{vinddk}\textbf{Curate}} \ {\footnotesize\color{inkmut}(merit-based lifecycle)}
\STATE \quad retire ops with non-positive leave-one-out marginal; promote shadows that raise dev agreement
\STATE {\color{ateal}\textbf{Audit}} \ {\footnotesize\color{inkmut}(measurement only, never a training signal)}
\STATE \quad report agreement against the locked test set
\end{algorithmic}
\end{algorithm}

\subsection{Co-evolution with the skill loop: Double Ratchet}

The Ratchet skill loop \citep{zhang2026ratchet}, adopted here unchanged, needs the one thing this setting does not supply: a grader for its training tasks. Co-evolution hands it the current best evolved verifier, and we call the paired system \emph{Double Ratchet}. The two loops alternate under a fixed front-loaded curriculum, verifier phases longest early when coverage is worst and shrinking as the verifier matures. In the skill phases the evolved verifier grades train attempts, and failed attempts become capsules whose error text feeds skill synthesis. The skill loop's held-out evaluation and rollback anchor stay pinned to the locked test set, which the evolved verifier never reads, so a corrupted verifier can slow skill learning but cannot redefine the evaluation metric. The comparison that matters is the \emph{reference} skill loop, identical machinery handed the ground-truth verifier (or best available rubric) for free.

\paragraph{Two ablations, one per safety ingredient.}
The \textbf{naive} arm disables the two anchor guards (fail-closed anchoring and the validity gate), plus skill-loop rollback on the skill side, with the detector lifecycle fully on, the collusion-prone configuration that unanchored agent-judge co-evolution implicitly runs. The \textbf{no-lifecycle} arm does the opposite, keeping the guards but disabling the birth gate, shadow tier, and merit-based retirement, so every synthesized detector enters active and nothing retires: the verifier-side analog of an unmanaged skill library.

\paragraph{An outer audit.}
Where no accurate verifier exists, rubric scores are references and not truth, so a final audit sits outside all loops: a stronger LLM compares each final output pairwise against the pre-evolution baseline, judged twice with positions swapped, a win counting only when both orders agree (Appendix~\ref{app:judge}). This mimics the human acceptance step a real deployment keeps, and it exposed a proxy-gaming failure the loops could not see. It is an audit, never a training signal.

\section{Experimental setup}

\paragraph{Task areas.}
\textbf{MBPP+} \citep{austin2021mbpp,liu2023evalplus}, Python function synthesis with hidden unit tests as the hard anchor, and \textbf{Spider~2.0-Snow} \citep{lei2025spider2}, enterprise text-to-SQL over real Snowflake warehouses anchored by official execution-result comparison, are the validation tasks: their exact ground truth makes both verifier validity and the reference-loop match measurable. \textbf{Report generation} is the target task, a deployment-style workload in which a pipeline writes analyst-report sections from audited evidence cards. No golden verifier exists for it; the available rubric score plus golden demonstration sections is a partial reference signal, not ground truth. Task cards are in Appendix~\ref{app:tasks}.

\paragraph{Hard subsets and splits.}
For MBPP+ and Spider~2.0 we screen every task with five independent frozen-solver samples, drop those passed on every sample (no lift room), and stratify the rest by screen pass rate. Report tasks derive from golden demonstrations and need no screening. Dev always holds ten items, five clear failures and five clean positives. The solver is a single direct LLM call with no tools, iteration, or scaffolding, so absolute numbers are not comparable to leaderboard agents. Appendix~\ref{app:setup} gives the splits, the anchors, and the Spider prompt grounding, a large shared effect applied identically to every arm.

\paragraph{Roles and models.}
All experiments use Claude models: every in-loop role (solver, skill synthesizers, teachers, judge ops) is Claude Opus~4.7, the sole exception being the report task's final judge, the stronger Claude Opus~4.8 kept outside all loops. A same-family teacher is not a circular self-grader: it compares candidate outputs against golden references on dev rather than judging unaided, and never sees the locked anchor. Skill loops run 100 rounds; co-evolution interleaves verifier phases of 15/8/5/2 rounds between four 25-round skill phases, matching the reference loop's skill budget. Every run is seeded and configuration-driven (Appendix~\ref{app:repro}).

\section{Results}

Two axes organize the evidence, taken in order. First \emph{verifier validity}, agreement with a locked reference no part of verifier evolution reads, which is the claim the guard ablations then localize. Then \emph{task score}, which we show cannot certify a verifier and therefore use only to ask whether an evolved one suffices in place of ground truth. Each run uses three seeds, reported as mean and sample standard deviation, with loop comparisons adding the seed-level difference and a 95\% bootstrap interval. Those means are coarse, so validity claims are also tested \emph{paired} over matched per-item verdicts: evolution rescores the same locked items, two rounds therefore differ only in the grader, and the test asks about that change of grader rather than any task-level population effect. Those matched pairs are clustered, three seeds over the same locked tasks, so exact McNemar over pooled pairs would be anti-conservative and the reported test is a sign-flip permutation at the task level. Per-round noise runs near five points, so we report peak and final-rounds values (mean of the last ten rounds) against each run's round-0 baseline; peak is a locked-set argmax and therefore optimistic, while the final-rounds column is fixed in advance and unselected, which is why the paired test is run at both.

\subsection{The evolved verifier beats the engineering it starts from}
\label{sec:validity}

On MBPP+ the evolved composition reaches $0.625\pm0.050$ held-out agreement against $0.417\pm0.058$ for the hand-authored seed composition it starts from, so the $+0.21$ is lift over static engineering rather than over nothing. It also beats the bare LLM judge op it contains ($0.55\pm0.04$), the natural baseline for a grader built without ground truth, and unlike that judge it is inspectable and diagnosable. The paired test makes the lift consistent but not significant: across 120 matched verdicts, evolution corrects 61 items against 36 broken at peak and 36 against 20 at the final rounds, with every seed net positive, yet clustering those verdicts by task leaves $p=0.12$ at peak and $p=0.21$ at the final rounds. Pooling them as independent would report $p=0.014$ and $p=0.044$: with 40 locked tasks and three seeds the design supports a consistent direction, not a calibrated rejection.

\begin{figure*}[t]
\centering
\includegraphics[width=\textwidth]{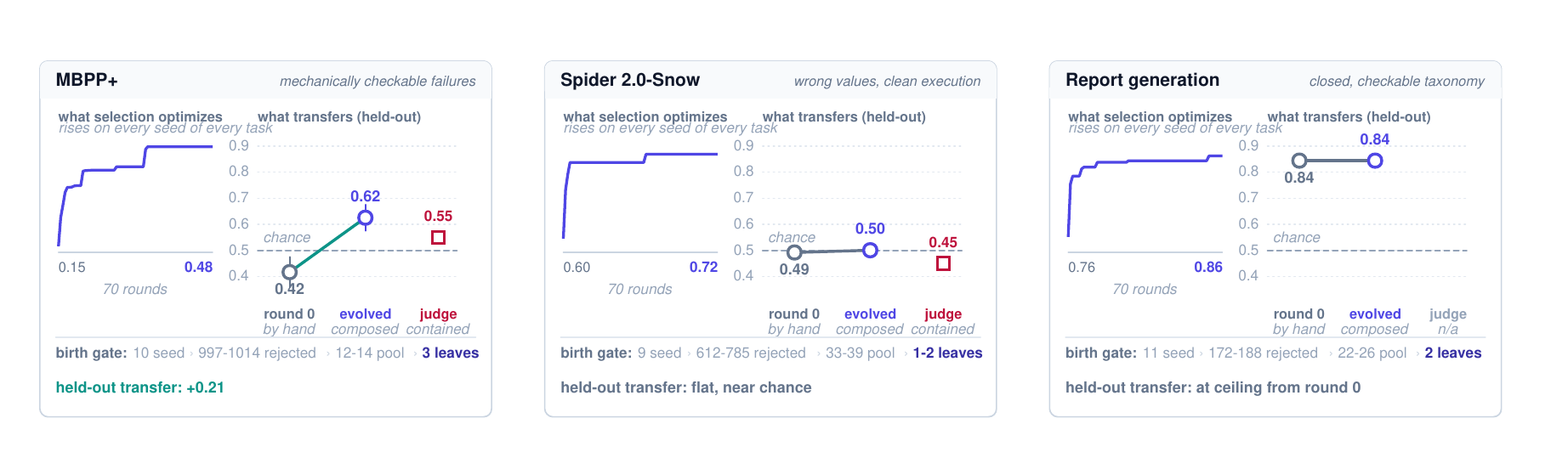}
\caption{Verifier validity, separating what selection sees from what it does not. \textbf{Left of each panel}: the selection objective over the first 70 of 100 rounds (three-seed mean). \textbf{Right}: held-out agreement with the locked anchor, from the hand-authored round-0 composition to the evolved peak (mean $\pm$ std, three seeds), against the bare LLM judge op contained in the composition and the chance line. \textbf{Bottom}: the birth-gate funnel, three-seed range.}
\label{fig:validity}
\end{figure*}

Figure~\ref{fig:validity} separates two quantities that are easy to confuse. What selection optimizes rises on every seed of every task and never degrades afterwards, which best-so-far selection guarantees and which is therefore no evidence of validity; only transfer to the locked set that selection never sees is informative. Transfer is also not monotone even though the objective is, since five of nine runs end below their own mid-run peak, the ordinary generalization gap of optimizing a ten-item proxy.

The other two tasks are not replications of the MBPP+ claim. Report generation starts at ceiling, because subset seeding already places a near-optimal composition in the round-0 candidate set, so it carries sufficiency and the safety ablations rather than lift. Spider sits near chance ($0.500\pm0.026$) despite a robust objective climb, and the paired test finds nothing there; \S\ref{sec:spectrum} shows this is that task's failure regime rather than a sample-size artifact, and \S\ref{sec:suff} why imperfect agreement does not doom the system.

The lifecycle does real work behind small expressions. Each pool starts from 9 to 11 hand-authored seed detectors, gap-driven synthesis then proposes continuously, and the birth gate rejects the overwhelming majority, leaving a working pool of a dozen to a few dozen from which anchored selection keeps one to three leaves. One Spider detector shows the whole path: the loop clustered train outputs that returned every column although the question named specific ones, authored the typed spec \texttt{forbids\_select\_star}, admitted it as a shadow because its birth evidence was an unlabeled gap, promoted it after composing it raised dev agreement from 0.71 to 0.81, and then selected it in place of the seed pool's grouping check (Appendix~\ref{app:artifacts}). Nor is the ten-item anchor fragile: replaying report selection against anchors of four, six, and eight items does not lower the mean held-out agreement and only widens its spread (Appendix~\ref{app:lifecycle}).

\begin{figure*}[t]
\centering
\includegraphics[width=\textwidth]{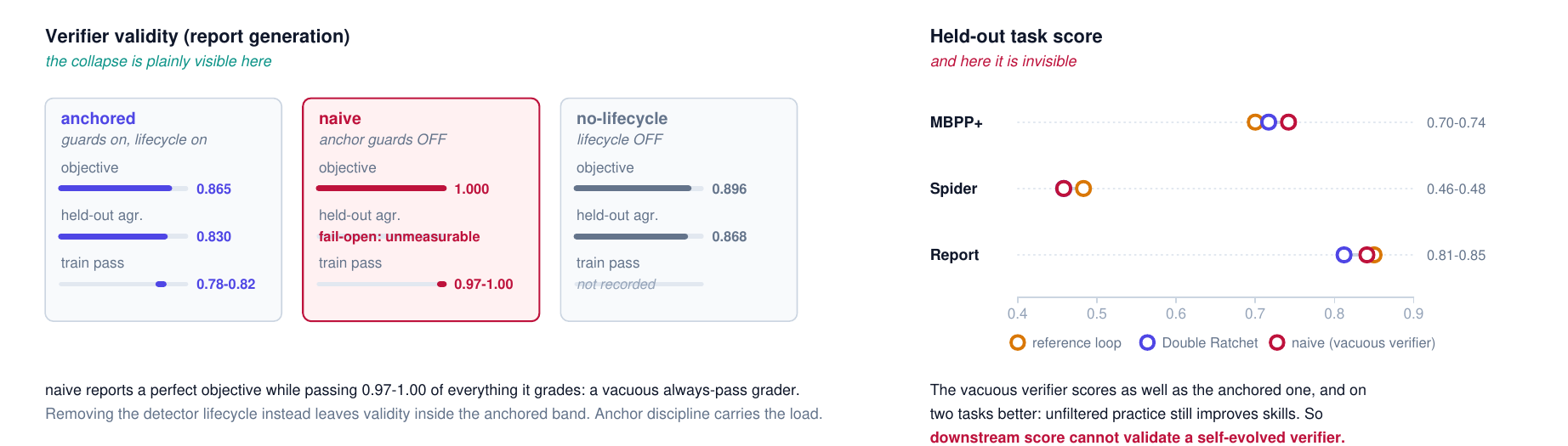}
\caption{The same three arms, measured two ways. \textbf{Left}: verifier validity on report generation, three seeds. The naive arm claims a perfect objective while passing 0.97--1.00 of everything, so its held-out agreement is fail-open and unmeasurable rather than merely low. \textbf{Right}: held-out task score for the reference loop, Double Ratchet, and the naive arm, which does not separate the three on any task.}
\label{fig:guards}
\end{figure*}

\subsection{Which guard carries the load, and why task score cannot tell}
\label{sec:guards}

The ablations are judged on verifier validity, not held-out task score, and Figure~\ref{fig:guards} shows why the distinction is not pedantic. On report generation, where the anchor is softest and collusion pressure highest, disabling the anchor guards collapses the verifier on three of three seeds: selection latches onto a detector that almost never fires, fail-open scoring reports a vacuous perfect objective, and the verifier passes 0.97--1.00 of everything it grades. Disabling the lifecycle collapses nothing: objectives and held-out agreement stay inside the anchored band, and although the pool grows, the extra detectors are inert, since one that is never selected never grades anything.

This inverts the emphasis of the skill-evolution literature. For skills, lifecycle management is the key finding; for verifiers under anchored selection, anchor discipline carries the safety load, so the verifier-side analog of library drift is \emph{anchor} drift, not pool drift. The asymmetry is structural: a junk skill gets routed into prompts and actively hurts, while a junk detector matters only if selection picks it, which the anchor prevents. The lifecycle still buys efficiency, and we keep it.

The collapsed arm carries a warning that outlives this paper. Its verifier is worthless, yet its co-loop scores as well as the anchored one and on two tasks slightly better (Figure~\ref{fig:guards}, right; Table~\ref{tab:main}). Neither number is wrong: an always-pass verifier reduces training to unfiltered practice, which on its own improves skills, and the measurement never passes through the verifier, so the task score survives what the grader does not. \textbf{Downstream task performance therefore cannot validate a self-evolved verifier}, and the experiment this literature reaches for first would here have certified a grader that passes everything. Validating the verifier requires a held-out reference it never reads, and any deployment that triages or gates on the verifier needs one anyway.

\subsection{Sufficiency: an evolved verifier substitutes for ground truth}
\label{sec:suff}

With validity measured independently, the remaining question is sufficiency: can a verifier evolved from ten reference examples drive a skill loop that ground truth would otherwise drive? It can substitute, which is a weaker claim than causing the gain: \S\ref{sec:guards} already showed retained lift does not isolate verifier accuracy. Double Ratchet retains 106\%, 110\%, and 88\% of the reference loop's lift on MBPP+, Spider, and Report (Table~\ref{tab:main}; curves in Figure~\ref{fig:curves}): within noise on both ground-truth tasks, where the $\Delta$peak intervals contain zero, and a small gap on the report task, whose reference is itself a partial rubric. That is why the independent judge in \S\ref{sec:hack} is the more informative read there.

Aggregates hide a task-level result. Both loops solve held-out problems the frozen solver \emph{never} solves, and on both hard-anchor tasks Double Ratchet converts \emph{more} of them than the ground-truth-driven loop (Table~\ref{tab:main}, \emph{Improved}). That settles how good a verifier must be to suffice here: not very. Spider's verifier agrees with ground truth only at $0.500\pm0.026$ yet its co-loop retains the full reference lift, because the verifier's role in training is \emph{directional}, deciding which attempts become failure capsules, whose concrete error text does not depend on the pass/fail label being right. The verifier-graded training pass rate stays in a realistic band (0.78--0.82 on Report, 0.41--0.83 on Spider), so the verifier is really grading. The reference loop and the co-loop, which never see each other, independently synthesize functionally matched skills on every task (Appendix~\ref{app:artifacts}).

The honest reading is that a skill loop is a \emph{forgiving} consumer of grades, and Spider's smallest absolute lift for both loops reflects one ceiling binding them alike: a single-call solver's residual failures mostly need live exploration that no guidance-text skill expresses. So the reasons to build a verifier well lie elsewhere, in triage, gating, and repairability, the last of which the next subsection puts under load.

\begin{table}[t]
\centering
\footnotesize
\caption{Held-out task score over three seeds on 40--48-item locked sets. \emph{reference}: the same skill loop handed ground truth (MBPP+, Spider) or the best available rubric (Report) for free. \emph{co}: Double Ratchet on its own evolved verifier. \emph{naive}$^\dagger$: the ablation with the anchor guards and skill-side rollback off, seed-matched to \emph{co}; its verifier is vacuous (Table~\ref{tab:ablation}) yet it scores as well as \emph{co}, which is why task score cannot certify a grader. \emph{Lift ret.}: co's peak lift over its round-0 baseline as a fraction of the reference loop's. \emph{Verifier agr.}: peak agreement with the locked anchor. \emph{Improved}: never-passed tasks solved at least once, or for Report the fraction of sections ending above round 0.}
\label{tab:main}
\setlength{\tabcolsep}{4pt}
\renewcommand{\arraystretch}{1.18}
\begin{tabular*}{\textwidth}{@{\extracolsep{\fill}} l ccc c c c cc @{}}
\toprule
 & \multicolumn{3}{c}{Peak held-out score} & lift & $\Delta$peak & verifier & \multicolumn{2}{c}{Improved} \\
\cmidrule(lr){2-4}\cmidrule(lr){8-9}
Task & reference & \pc co & naive$^\dagger$ & ret. & (95\% CI) & agr. & reference & \pc co \\
\midrule
MBPP+  & 0.700$\pm$.025 & \pc\textbf{0.717$\pm$.038} & 0.742$\pm$.014 & 106\% & $+.02\,[-.03,.06]$ & 0.62 & 16/23 & \pc\textbf{19/23} \\
Spider & 0.483$\pm$.038 & \pc\textbf{0.458$\pm$.038} & 0.458$\pm$.029 & 110\% & $-.03\,[-.08,.03]$ & 0.50 & 4/12  & \pc\textbf{6/12} \\
Report & 0.850$\pm$.010 & \pc\textbf{0.812$\pm$.006} & 0.841$\pm$.003 & 88\%  & $-.04\,[-.05,-.03]$ & 0.84 & 99\%  & \pc\textbf{99\%} \\
\bottomrule
\end{tabular*}
\end{table}

\subsection{A reward-hacking episode, and auditing the auditor}
\label{sec:hack}

The report task produced this paper's clearest evidence that an inspectable verifier is worth having. The rubric's metric-discipline dimension counts inline evidence tags, and the first skill runs raised the rubric score by $+0.26$ partly by gaming exactly that: evolved skills wrote the tag \emph{in place of} the number, leaving about 30\% of tags at peak rounds with no value beside them, and invented confident forecasts to satisfy a style dimension. This is textbook proxy gaming, and no in-loop signal saw it. The independent judge did, preferring the pre-evolution baseline in 88\% of decided pairs and describing unrendered placeholders and fabricated conviction.

The repair was one coupled change: a vocabulary-aware value-erasure check added to the capsule gate, where a tag whose registered metric has a numeric value but no numeral nearby is a defect, plus rewritten failure hints that teach value-plus-tag, the hints being how a detector's verdict reaches skill synthesis. We claim the pair rather than the detector alone, since separating them needs reruns we did not do. The rerun cut erased tags from roughly 30\% to about 1\% while the rubric score \emph{rose} to $0.850\pm0.010$, trading nothing away. Reading the verifier is what localized the defect to a nameable check, an intervention a learned scalar does not admit.

The audit then needed auditing, a concrete instance of the scalable-oversight problem. The pipeline requires raw evidence tags in its output, a format contract the original generic rubric did not know, so its loss rationales blamed that syntax and the repair barely moved the win rate, from 0.122 to 0.126. A task-aware rubric stating the contract, while still treating valueless tags and fabricated figures as defects, re-judged the same stored pairs. Table~\ref{tab:judge} separates the two layers: down the task-aware column the erasure fix moves the win rate from 0.515 to 0.770, a content change the contract makes visible, while each row's generic-to-task-aware jump measures the generic judge's convention-blindness. That rubric was written after reading the generic judge's rationales, so it diagnoses the audit rather than independently confirming the repair; its pre-repair cell is the control. The lesson runs both directions: self-evolved verifiers need an independent outer audit, and the audit needs the task contract.

\begin{table}[t]
\centering
\footnotesize
\caption{Final-judge win rate of evolved report outputs over their own pre-evolution baselines, over decided pairs; each pair is double-judged with positions swapped and a win requires both orders to agree (protocol and counts in Appendix~\ref{app:judge}). Rows isolate the one-detector erasure fix, columns the judge's convention-blindness: only the task-aware judge registers the fix ($+0.255$).}
\label{tab:judge}
\setlength{\tabcolsep}{5pt}
\renewcommand{\arraystretch}{1.2}
\begin{tabular*}{\textwidth}{@{\extracolsep{\fill}} l cc c >{\raggedright\arraybackslash}p{5.3cm} @{}}
\toprule
 & \multicolumn{2}{c}{Outer judge} & effect of & \\
\cmidrule(lr){2-3}
Report loop state & generic & task-aware & the rubric & What the row shows \\
\midrule
pre-repair (proxy gamed)   & 0.122 & 0.515 & $+0.393$ & only the aware judge sees past the syntax \\
post-repair (erasure fixed) & 0.126 & \pc\textbf{0.770} & $+0.644$ & content gain the generic judge cannot see \\
\midrule
\emph{effect of the fix} & $+0.004$ & $+0.255$ & & the generic judge penalizes the required format \\
\bottomrule
\end{tabular*}
\end{table}

\subsection{The detectability spectrum}
\label{sec:spectrum}

One variable organizes every result above: how mechanically detectable the solver's failures are. MBPP+ failures are crashes and wrong outputs that cheap detectors see directly, and held-out agreement lifts $+0.21$. Report generation derives its anchor from golden demonstrations through a closed, checkable failure taxonomy, and transfer reaches 0.84. Spider~2.0 gives an in-task controlled contrast: under our first prompt most failures were compile errors and the evolved verifier reached 0.85 held-out agreement. Grounding the prompt with typed schemas then raised the solver's baseline by half (0.211 to 0.329) and left \emph{residual} failures that are semantically wrong values under clean execution, invisible to deterministic detectors, so agreement fell to $0.500\pm0.026$ even as the objective still climbed. Same method, same task, different failure distribution. That regime shifts the burden onto judge ops and human audits, and sharper solvers leave more of it: verifier evolution buys the most where failures are checkable, a scoping condition rather than an implementation limit.

\section{Discussion and conclusion}

Evolution expands coverage but never creates ground truth. With no anchor the loop optimizes noise, so anchor \emph{quality} matters as much as its existence, and hardening soft anchors into checkable ones pays most. The open frontier is outputs that execute cleanly yet are semantically wrong. This is a mechanism study, not a scaling result: three task families, one model family, rounds in the low hundreds, held-out sets of tens of items, and one stronger model as outer judge rather than a panel or human study.

An agent lacking a verifier can evolve one under the discipline that stabilizes its skills, with safety engineered rather than assumed. Three commitments carried our results. \emph{Anchor what you cannot manufacture}: tie selection to a small reference set the loop cannot edit, never to the agent's own score. \emph{Keep the verifier legible}: mostly deterministic detectors fail differently from the model they grade, and one named check repaired a live case of proxy gaming. \emph{Audit the auditor}: an outer judge is necessary but not sufficient, and ours was wrong until given the task contract. Never validate a self-evolved verifier by the agent's performance: a grader that passes everything passes that test too. Who verifies the verifier? An anchor it predicts but never sees, and a judge outside every loop.

\bibliographystyle{unsrtnat}
\bibliography{references}

\appendix

\section{Task descriptions}
\label{app:tasks}

\paragraph{MBPP+ (validation task; anchor: hidden unit tests).}
A one-paragraph natural-language programming problem plus one example call; the solver writes a Python function. The anchor is the EvalPlus hidden test suite \citep{liu2023evalplus}, which adds many more tests per problem than the original MBPP and repairs faulty ones, giving reliable pass/fail. Typical failures are crashes, mishandled edge cases, printing instead of returning, and misread specifications, which are the failure classes the detector pool checks.

\paragraph{Spider 2.0-Snow (validation task; anchor: execution match).}
An analyst question over a real hosted Snowflake warehouse: production-scale databases with hundreds of tables, cryptic columns, sharded table families, and dialect-specific SQL. The solver writes one query from the question, a rendered schema catalog, and optional domain notes. The anchor executes the candidate and compares the fetched table to the stored gold result with the official value-based comparator, ignoring column names and order. Typical failures are unresolvable identifiers, type errors on date-like columns, and, hardest, queries that run cleanly but compute the wrong thing. Our single-call solver passes about a third of screened tasks.

\paragraph{Report generation (target task; no accurate verifier exists).}
Write one \emph{section} of an analyst report (for example, ``value-chain map: control points from ASIC to optical engine'') in markdown, given a section title and purpose, a style contract fixing units, reference dates, and accounting rules, a set of audited evidence cards, and a metric vocabulary registering the section's key figures. An evidence card carries an id, a one-sentence claim, a verbatim source quote, source attribution and date, a stance, and an auditor's credibility verdict. The section must ground every claim in the cards, state each key figure with the inline tag binding it to the metric vocabulary, hedge weak or disputed evidence, and include required structure such as a summary line. No verifier exists because section quality is holistic: even an expert author cannot reduce ``a good section'' to a checkable rule. The rubric scores lexical and structural proxies of the requirements above and is treated as reference signal only.

\section{Setup details}
\label{app:setup}

\paragraph{Splits and screening.}
For MBPP+ and Spider~2.0 we screen every task with five independent samples of the frozen single-call solver, graded by the official anchor, drop tasks the solver always passes, stratify the rest across pass levels, then split 60/40 into train and test. The resulting train/dev/test sizes are 60/10/40 for MBPP+, 59/10/40 for Spider~2.0, and 73/10/48 for report generation. The dev set is ten items, five clear failures and five clean positives (for report generation, five golden-quality positives and five class-stratified checkable failures), so the verifier is scored on both classes.

\paragraph{Anchors.}
\emph{MBPP+}: hidden unit tests. \emph{Spider~2.0}: execute on the live warehouse and compare the fetched table to the stored gold result with the official comparator; we reproduce the evaluator's CSV round-trip of the fetched frame, without which even gold queries fail on dtype. \emph{Report generation}: the verifier loop's anchor is a frozen demo-delta soft label, where a teacher compares the frozen-agent report against the golden section under a closed seven-class miss taxonomy and passes only if at most one miss class is present. Each of the 48 held-out sections contributes two locked items, the frozen-agent report (which the strict anchor almost always fails) and the golden section (which passes by definition), giving 96 items split 54 pass and 42 fail, so a degenerate always-pass verifier scores 0.56. The skill and co-loops are evaluated by the official rubric evaluator (0--100 normalized), whose dimensions cover evidence grounding, metric-tag discipline, claim calibration, structure, and style.

\paragraph{Spider~2.0 prompt grounding.}
The generation prompt grounds the solver with a compact typed schema catalog (columns as name:TYPE, sharded tables collapsed, sample values on date-like columns) and Snowflake-specific rules. We report this because it is a large shared effect we control for, not a contribution: grounding raises the frozen baseline by half on the screened pool (0.211 to 0.329) but is applied identically to every arm, so all reported lifts sit on top of it. It also sets up the detectability contrast of \S\ref{sec:spectrum}, since grounding pushes residual failures from compile errors into clean-executing wrong values.

\section{Expression grammar, selection, and lifecycle constants}
\label{app:lang}

Writing $v(e) \in \{1,0,\bot\}$ for the verdict of expression $e$ on $(t,y)$, with $1$ denoting \textsc{drawback} and $\bot$ \textsc{abstain}, the expression grammar of \S\ref{sec:ops} is
\begin{equation}
e \mathrel{::=} o
  \;\big|\; \textstyle\bigvee_{i \leq k} e_i
  \;\big|\; \textstyle\bigwedge_{i \leq k} e_i
  \;\big|\; \neg e
  \;\big|\; \textstyle\big[\sum_{i \leq k} v(e_i) \geq K\big]
  \;\big|\; \textstyle\big[\sum_{i \leq k} w_i v(e_i) \geq \theta\big],
\end{equation}
where $|$ separates alternative node forms: a leaf op, a disjunction, a conjunction, a negation, an unweighted $K$-of-$k$ vote, or a weighted vote against threshold $\theta$.

The objective multiplies dev agreement by consensus agreement on unlabeled train outputs (exponent 1.0 for MBPP+ and Spider, 0.25 for report generation, where detector consensus is a weaker regularizer), minus a small complexity penalty. Dev agreement is the recall-weighted mean of the two per-class recalls over the dev items a candidate opines on, with the \textsc{fail} class at weight 2 and \textsc{clean} at weight 1. Train consensus is reliability-weighted, each opining op's vote scaled by $1 + 10\max(0, m_o)$ for that op's anchored leave-one-out marginal $m_o$, so anchor-proven ops lead the vote and ops with no anchored evidence contribute at unit weight; this keeps a bloc of cheaply synthesized correlated detectors from defining the pseudo-label the regularizer rewards agreeing with. The complexity penalty charges 0.001 per op beyond a floor of three and a further 0.002 per LLM-calling op, which is why selected expressions stay small and mostly deterministic. Fail-closed anchoring makes candidates without a usable dev opinion unselectable, and a validity gate drops all-pass, all-fail, and all-abstain candidates. The birth gate requires a new op to fire on at least half its cluster and stay clean on known-good outputs; ops born from anchored misses enter active, unlabeled-gap ops enter shadow and are promoted only on demonstrated anchor agreement; per-op fitness is the leave-one-out marginal of the incumbent expression and non-positive ops retire after a two-round grace. The \textbf{naive} ablation disables exactly the two anchor guards (and skill-loop rollback); the \textbf{no-lifecycle} ablation disables exactly the birth gate, shadow tier, and retirement.

\section{Final-judge protocol and rubrics}
\label{app:judge}

For each of the 48 held-out sections we pair the stored round-0 report against the stored peak-round report, and separately against the final-round report, for every arm and seed. Pre-repair this is 3 arms $\times$ 3 seeds $\times$ 2 rounds $\times$ 48 sections $=$ 864 pairs; the post-repair rerun adds a fourth seed on the two co-loop arms, giving 1{,}056. Each pair is judged twice with positions swapped, and a win counts only when both orders agree, otherwise it is a tie. The generic rubric asks a senior-editor question about holistic reader value with no knowledge of the pipeline's conventions. The task-aware rubric additionally states the format contract: evidence citations and inline metric tags are required output that renders downstream, a tag with no value nearby is a defect, and fabricated figures are disqualifying. Both rubrics judge the same stored pairs, so the four cells of Table~\ref{tab:judge} differ only in the repair pass (rows) and the rubric (columns). The decided-pair counts behind the task-aware column are 271 of 526 pre-repair and 435 of 565 post-repair.

\section{What evolution actually produced}
\label{app:artifacts}

\paragraph{Final verifiers, verbatim.}
On MBPP+, \texttt{(any spec\_mismatch crash returns\_not\_print\_only)}: flag a solution if the judge finds a specification mismatch, or it crashes on probe inputs, or it prints instead of returning. On Spider~2.0, \texttt{(any missing\_group\_by spec\_sql\_mismatch)}: flag a query if it aggregates without grouping, or the judge finds it does not answer the question; this form was selected independently on two of three seeds. On report generation, \texttt{(any ledger\_metric\_uncovered low\_status\_card\_overclaimed)}: flag a section if a registered key figure never appears, or a weak-credibility card is cited without hedging; the other two seeds swap the first leaf for an uncited-claims detector.

\paragraph{One detector through the whole lifecycle.}
On the Spider seed that ends at \texttt{(any syn\_forbids\_select\_star\_when\_columns spec\_sql\_mismatch)}, the loop clustered four unlabeled train outputs that returned every column although the question implied specific ones, and authored the typed spec \texttt{\{kind: static, check: forbids\_select\_star, detects: select\_star\_specific\_columns\}}. Because its birth evidence was an unlabeled gap rather than an anchored miss, it entered as a \emph{shadow}: it ran and recorded verdicts but could not be selected. It was promoted in the same round, but only after composing it with the incumbent raised dev agreement from 0.71 to 0.81, and selection then adopted it, replacing the grouping check the seed pool had contributed. Two of the co-loop seeds on this task converge on a different synthesized detector (\texttt{syn\_top\_n\_missing\_limit}) and the third on the seed \texttt{select\_star} check, so the leaf is not a fixed artifact of the pool.

\paragraph{Grading one task: hidden tests against the evolved verifier.}
For ``write a function to find the ratio of zeroes in an array of integers'', the solver returned \texttt{zeros / non\_zeros}: the code runs cleanly and returns a number, but computes the ratio to non-zeros where the specification wants the ratio to all elements, so only the meaning is wrong. The hidden unit tests raise \texttt{AssertionError} and return \textsc{fail}, a reliable verdict with no indication of which requirement was misread. The evolved composition returns \textsc{fail} as well, from \texttt{crash}: clean, \texttt{returns\_not\_print\_only}: clean, \texttt{spec\_mismatch}: \textsc{drawback}, and so \emph{names} the failure class. That named drawback is what turns the attempt into a usable failure capsule for skill synthesis; the anchor's bare verdict could not have driven the same repair.

\paragraph{Same lesson, two loops.}
Each pair below was synthesized independently, one by the reference loop under ground-truth grading and one under the evolved verifier, and neither loop saw the other. \emph{MBPP+}: reference \texttt{function-name-from-assert}, ``the harness imports the symbol literally as written in the assert; lift the identifier verbatim including case, the prose is only a hint''; co \texttt{copy\_callable\_identifier\_from\_assert}, ``the grader imports the name verbatim from the assert; reproduce that exact spelling letter for letter, including apparent typos''. \emph{Spider~2.0}: reference \texttt{sql-empty-result-diagnosis}, ``a 0-row result is almost never correct when the gold has rows; incrementally relax predicates to locate which condition eliminates all rows''; co \texttt{sql-empty-result-sanity-probe}, ``an empty result almost always means one predicate matched nothing; validate each CTE's row count in isolation before trusting the join''. \emph{Report}: reference \texttt{metric-tag-annotates-not-replaces}, ``the scorer counts a figure only when a literal number sits adjacent to the brace tag''; co \texttt{metric-tag-inline-after-number}, ``the tag annotates the value; it never replaces the value, and a bare card cite does not satisfy it''.

\section{Lifecycle statistics and full per-loop scores}
\label{app:lifecycle}

\begin{table}[h]
\centering
\footnotesize
\caption{Verifier-validity ablations on report generation, final-rounds values over three seeds, each arm flipping one guard family (\gon\ on, \goff\ off). Removing the \emph{anchor} guards collapses the verifier into a vacuous always-pass grader on every seed; removing the detector \emph{lifecycle} does not. \emph{Train pass}: fraction of training attempts passed (\emph{n/r}: not recorded for this arm). The collapse is invisible in held-out task score (Table~\ref{tab:main}, \emph{naive}). Tinted row: our configuration.}
\label{tab:ablation}
\setlength{\tabcolsep}{5pt}
\renewcommand{\arraystretch}{1.18}
\begin{tabular*}{\textwidth}{@{\extracolsep{\fill}} l cc ccc l @{}}
\toprule
 & \multicolumn{2}{c}{Guards} & \multicolumn{3}{c}{Verifier validity} & \\
\cmidrule(lr){2-3}\cmidrule(lr){4-6}
Arm & anchor & lifecycle & objective & held-out agr. & train pass & Outcome \\
\midrule
\pc\textbf{anchored} & \pc\gon & \pc\gon & \pc 0.865$\pm$0.002 & \pc\textbf{0.830$\pm$0.012} & \pc .78--.82 & \pc valid, composes \\
naive        & \goff & \gon  & 1.000 \emph{(vacuous)} & \emph{fail-open} & .97--1.0 & {\color{frose}\textbf{collapses}} to always-pass \\
no-lifecycle & \gon  & \goff & 0.896$\pm$0.072 & 0.868$\pm$0.061 & \emph{n/r} & no collapse \\
\bottomrule
\end{tabular*}
\end{table}

\begin{figure*}[h]
\centering
\includegraphics[width=\textwidth]{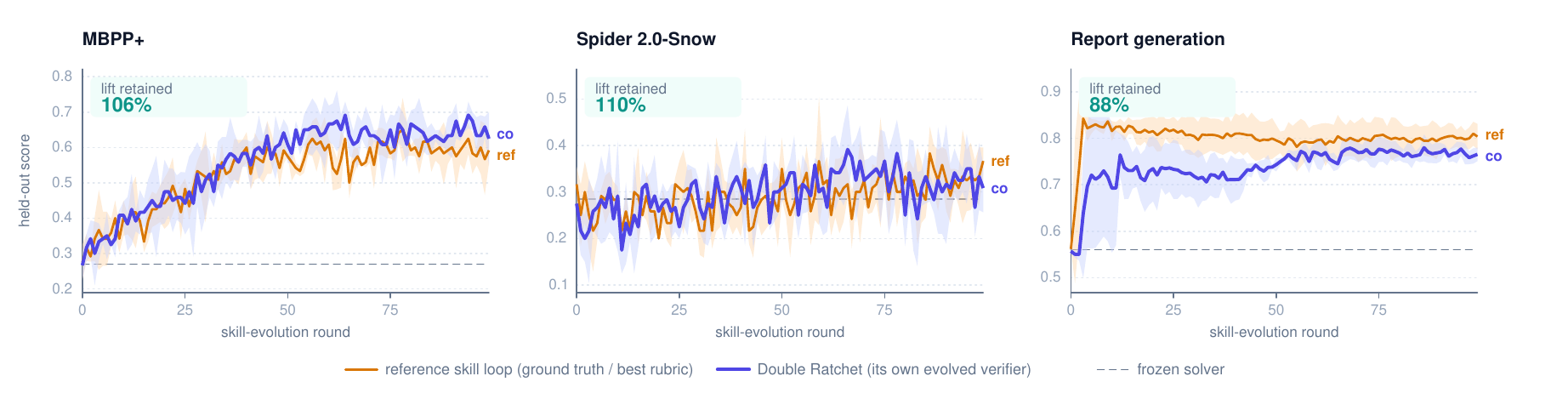}
\caption{Held-out learning curves (mean $\pm$ std, three seeds): the reference skill loop on ground truth or the best available rubric, against Double Ratchet on its own evolved verifier. Callout: co-evolution's lift retention as a fraction of the reference loop's. Held-out evaluation never passes through the evolved verifier.}
\label{fig:curves}
\end{figure*}

Figure~\ref{fig:curves} plots the held-out learning curves behind Table~\ref{tab:main}, Table~\ref{tab:oplife} recaps what the detector lifecycle did in the anchored arms, and Table~\ref{tab:perloop} gives the complete held-out init, peak, and final-rounds scores. Frozen single-call baselines are approximately 0.27 (MBPP+), 0.285 (Spider~2.0), and 0.56 (report generation).

\paragraph{Skill-side behavior.}
Synthesis is prolific at roughly 1.5 skills per round, the bounded bank forces contribution-based eviction of more than half of everything created, and rollback fires rarely. The report skill loop is the outlier: its targeted failure capsules produce skills effective enough that a handful suffice, with one seed holding its plateau on a bank of six.

\begin{table}[h]
\centering
\footnotesize
\caption{Detector lifecycle, anchored arm over 100 rounds. Three-seed min--max; a bare number means all seeds agree. Anchored selection prefers precise small compositions over detector accumulation.}
\label{tab:oplife}
\setlength{\tabcolsep}{5pt}
\renewcommand{\arraystretch}{1.15}
\begin{tabular*}{\textwidth}{@{\extracolsep{\fill}} l ccccc @{}}
\toprule
Task & seed detectors & born & birth-gate rejections & final pool & leaves in verifier \\
\midrule
MBPP+  & 10 & 2--4   & 997--1014 & 12--14 & 3 \\
Spider & 9  & 24--30 & 612--785  & 33--39 & 1--2 \\
Report & 11 & 11--15 & 172--188  & 22--26 & 2 \\
\bottomrule
\end{tabular*}
\end{table}

\begin{table}[h]
\centering
\footnotesize
\caption{Held-out evaluation per loop, mean $\pm$ std over three seeds; the \emph{naive} rows use the same three seeds as \emph{co} on each task. Final rounds: mean of each run's last ten rounds. Tinted: our configuration.}
\label{tab:perloop}
\setlength{\tabcolsep}{5pt}
\renewcommand{\arraystretch}{1.15}
\begin{tabular*}{\textwidth}{@{\extracolsep{\fill}} ll ccc @{}}
\toprule
Task & Loop & init & peak & final rounds \\
\midrule
MBPP+  & reference & 0.275$\pm$0.050 & 0.700$\pm$0.025 & 0.592$\pm$0.039 \\
       & \pc co    & \pc 0.267$\pm$0.038 & \pc\textbf{0.717$\pm$0.038} & \pc\textbf{0.652$\pm$0.034} \\
       & naive     & 0.317$\pm$0.058 & 0.742$\pm$0.014 & 0.632$\pm$0.023 \\
\midrule
Spider & reference & 0.317$\pm$0.038 & 0.483$\pm$0.038 & 0.327$\pm$0.010 \\
       & \pc co    & \pc 0.275$\pm$0.043 & \pc\textbf{0.458$\pm$0.038} & \pc\textbf{0.323$\pm$0.038} \\
       & naive     & 0.292$\pm$0.052 & 0.458$\pm$0.029 & 0.343$\pm$0.028 \\
\midrule
Report & reference & 0.562$\pm$0.001 & 0.850$\pm$0.010 & 0.802$\pm$0.024 \\
       & \pc co    & \pc 0.557$\pm$0.006 & \pc\textbf{0.812$\pm$0.006} & \pc\textbf{0.767$\pm$0.015} \\
       & naive     & 0.559$\pm$0.003 & 0.841$\pm$0.003 & 0.811$\pm$0.015 \\
\bottomrule
\end{tabular*}
\end{table}

\paragraph{Two further scope limits.}
Guidance-text skills cannot substitute for capabilities that need interaction, which is the ceiling both loops hit on Spider~2.0 (\S\ref{sec:suff}). The op type system also admits failure-anticipating detectors in the spirit of metamorphic testing \citep{chen1998metamorphic}, which perturb a task and check that the output changes as it should; these are implemented but no selected verifier used one, so we make no claim about them.

\paragraph{Naive co-loops are not a free pass.}
Naive co-loops match the anchored arm on held-out evaluation because their measurement anchor is untouched, but their verifiers are degenerate on every task, not only on report generation (\S\ref{sec:guards}): on Spider~2.0 the naive objective barely moves on two of three seeds.

\paragraph{Skill-loop robustness to verifier quality.}
Per-seed evidence for the directional-signal argument of \S\ref{sec:suff}: two of three Spider co-evolution seeds selected a weak synthesized single-detector verifier (train agreement 0.41--0.44) yet reached peak evaluations of 0.45--0.50, indistinguishable from the sibling seed with a two-detector verifier.

\paragraph{Anchor size.}
Ten dev items is a design choice about what sparse anchoring realistically supplies, not a tuned setting. The replay behind the \S\ref{sec:validity} claim takes report-generation selection offline, draws 200 stratified subsamples per anchor size from the same ten items, preserves the fail/clean balance within each draw, and rescores every candidate expression the live run actually considered under the reduced anchor. No new generation and no new solver calls are involved, so the comparison isolates the anchor's contribution to selection from run-to-run variance. Held-out agreement is $0.871$, $0.882$, and $0.874$ at four, six, and eight anchored items against $0.854$ at ten. Subsample standard deviations grow as the anchor shrinks (0.063 at four items against 0.041 at eight), the expected cost of a smaller anchor: any single small anchor is a lottery, while the mean over draws shows selection is not relying on the marginal item.

\section{Reproducibility}
\label{app:repro}

No GPUs are involved, since every model call goes to a hosted API, so a run needs only a CPU host (ours: 48-core x86-64, 186\,GB RAM, Ubuntu~24.04, Python~3.12, with \texttt{pydantic}, \texttt{pyyaml}, \texttt{boto3}, and for Spider~2.0 \texttt{snowflake-connector-python}), and wall-clock time is dominated by API latency. Each run writes its round-by-round history, detector pool, skill bank, and judge outputs as structured JSON and SQLite files, and every number in this paper is read directly from those files; the per-run databases store all 52{,}600 locked verdicts of the reported verifier runs with their per-detector traces, which is what makes the paired tests recomputable without rerunning any loop. Every LLM role (solver, teacher, composer, skill synthesizer, judge ops, final judge) is bound per call from a role-to-model table, so repeating any experiment under a different model or a mixed assignment is a configuration change rather than a code change; we hold that table uniform here to keep the comparisons clean. Spider~2.0 requires credentialed access to the benchmark's hosted warehouse; all other experiments are self-contained.

\end{document}